\documentclass[10pt,twocolumn,letterpaper]{article}

\usepackage{cvpr}              

\usepackage[accsupp]{axessibility} 

\usepackage[dvipsnames]{xcolor}
\usepackage{graphicx}
\usepackage{amsmath}
\usepackage{amssymb}
\usepackage{booktabs}

\usepackage{times}
\usepackage{epsfig}
\usepackage{graphics} 
\usepackage{mathptmx} 
\usepackage{amsfonts} 
\usepackage{helvet}

\usepackage{url}
\usepackage{rotating}
\usepackage{xspace}
\usepackage{booktabs}       
\usepackage{nccmath}
\usepackage{nicefrac}       
\usepackage{microtype}      
\usepackage{comment}		
\usepackage{adjustbox}		
\usepackage{tabu}
\usepackage{calc}			
\usepackage{wrapfig}
\usepackage{tabularx}
\usepackage{array}
\usepackage{makecell}

\usepackage{multirow}
\usepackage{float}
\usepackage[hang,flushmargin]{footmisc}
\usepackage[neverdecrease]{paralist}
\usepackage{algorithm}
\usepackage{algorithmicx}
\usepackage{algpseudocode}

\usepackage{calc}
\usepackage{color}
\usepackage{colortbl}
\usepackage{pifont}
\usepackage{animate}
\usepackage{nicefrac}       
\usepackage{xfrac}

\definecolor{cvprblue}{rgb}{0.21,0.49,0.74}
\usepackage[pagebackref,breaklinks,colorlinks,citecolor=cvprblue]{hyperref}

\usepackage[capitalize]{cleveref}
\crefname{section}{Sec.}{Secs.}
\Crefname{section}{Section}{Sections}
\Crefname{table}{Table}{Tables}
\crefname{table}{Tab.}{Tabs.}

\definecolor{CuGray}{gray}{0.9}
\newcolumntype{g}{>{\columncolor{CuGray}}c}

\makeatletter
\DeclareRobustCommand\onedot{\futurelet\@let@token\@onedot}
\def\@onedot{\ifx\@let@token.\else.\null\fi\xspace}

\newcommand{\dashrule}[1][black]{%
  \color{#1}\rule[\dimexpr.5ex-.2pt]{4pt}{.4pt}\xleaders\hbox{\rule{4pt}{0pt}\rule[\dimexpr.5ex-.2pt]{4pt}{.4pt}}\hfill\kern0pt%
}

\def\eg{\emph{e.g}\onedot} 
\def\ie{\emph{i.e}\onedot}

\def\paperID{8} 
\def\confName{CVPR Workshop}
\def\confYear{2024}

\title{AAPL: Adding Attributes to Prompt Learning for Vision-Language Models}
\author{Gahyeon Kim \footnotemark \qquad
Sohee Kim \footnotemark[\value{footnote}] \qquad
Seokju Lee \footnotemark \\
Korea Institute of Energy Technology (KENTECH)\\
\tt\small {\{gahyeon, soheekim, slee\}@kentech.ac.kr}
}

\begin{document}
%
%
%
%
%
%
%

\newcommand{\ba}{{\mathbf{a}}}
\newcommand{\bb}{{\mathbf{b}}}
\newcommand{\bc}{{\mathbf{c}}}
\newcommand{\bd}{{\mathbf{d}}}
\newcommand{\bolde}{{\mathbf{e}}}
\newcommand{\boldf}{{\mathbf{f}}}
\newcommand{\bg}{{\mathbf{g}}}
\newcommand{\bh}{{\mathbf{h}}}
\newcommand{\bi}{{\mathbf{i}}}
\newcommand{\bj}{{\mathbf{j}}}
\newcommand{\bk}{{\mathbf{k}}}
\newcommand{\bl}{{\mathbf{l}}}
\newcommand{\bm}{{\mathbf{m}}}
\newcommand{\bn}{{\mathbf{n}}}
\newcommand{\bo}{{\mathbf{o}}}
\newcommand{\bp}{{\mathbf{p}}}
\newcommand{\bq}{{\mathbf{q}}}
\newcommand{\br}{{\mathbf{r}}}
\newcommand{\bs}{{\mathbf{s}}}
\newcommand{\bt}{{\mathbf{t}}}
\newcommand{\bu}{{\mathbf{u}}}
\newcommand{\bv}{{\mathbf{v}}}
\newcommand{\bw}{{\mathbf{w}}}
\newcommand{\bx}{{\mathbf{x}}}
\newcommand{\by}{{\mathbf{y}}}
\newcommand{\bz}{{\mathbf{z}}}

\newcommand{\bA}{\mathbf{A}}
\newcommand{\bB}{\mathbf{B}}
\newcommand{\bC}{\mathbf{C}}
\newcommand{\bD}{\mathbf{D}}
\newcommand{\bE}{\mathbf{E}}
\newcommand{\bF}{\mathbf{F}}
\newcommand{\bG}{\mathbf{G}}
\newcommand{\bH}{\mathbf{H}}
\newcommand{\bI}{\mathbf{I}}
\newcommand{\bJ}{\mathbf{J}}
\newcommand{\bK}{\mathbf{K}}
\newcommand{\bL}{\mathbf{L}}
\newcommand{\bM}{\mathbf{M}}
\newcommand{\bN}{\mathbf{N}}
\newcommand{\bO}{\mathbf{O}}
\newcommand{\bP}{\mathbf{P}}
\newcommand{\bQ}{\mathbf{Q}}
\newcommand{\bR}{\mathbf{R}}
\newcommand{\bS}{\mathbf{S}}
\newcommand{\bT}{\mathbf{T}}
\newcommand{\bU}{\mathbf{U}}
\newcommand{\bV}{\mathbf{V}}
\newcommand{\bW}{\mathbf{W}}
\newcommand{\bX}{\mathbf{X}}
\newcommand{\bY}{\mathbf{Y}}
\newcommand{\bZ}{\mathbf{Z}}

\newcommand{\calA}{{\mathcal{A}}}
\newcommand{\calB}{{\mathcal{B}}}
\newcommand{\calC}{{\mathcal{C}}}
\newcommand{\calD}{{\mathcal{D}}}
\newcommand{\calE}{{\mathcal{E}}}
\newcommand{\calF}{{\mathcal{F}}}
\newcommand{\calG}{{\mathcal{G}}}
\newcommand{\calH}{{\mathcal{H}}}
\newcommand{\calI}{{\mathcal{I}}}
\newcommand{\calJ}{{\mathcal{J}}}
\newcommand{\calK}{{\mathcal{K}}}
\newcommand{\calL}{{\mathcal{L}}}
\newcommand{\calM}{{\mathcal{M}}}
\newcommand{\calN}{{\mathcal{N}}}
\newcommand{\calO}{{\mathcal{O}}}
\newcommand{\calP}{{\mathcal{P}}}
\newcommand{\calQ}{{\mathcal{Q}}}
\newcommand{\calR}{{\mathcal{R}}}
\newcommand{\calS}{{\mathcal{S}}}
\newcommand{\calT}{{\mathcal{T}}}
\newcommand{\calU}{{\mathcal{U}}}
\newcommand{\calV}{{\mathcal{V}}}
\newcommand{\calW}{{\mathcal{W}}}
\newcommand{\calX}{{\mathcal{X}}}
\newcommand{\calY}{{\mathcal{Y}}}
\newcommand{\calZ}{{\mathcal{Z}}}
\newcommand{\calbX}{\mbox{\boldmath $\mathcal{X}$}}
\newcommand{\calbY}{\mbox{\boldmath $\mathcal{Y}$}}

\newcommand{\bcalA}{\mbox{\boldmath $\calA$}}
\newcommand{\bcalB}{\mbox{\boldmath $\calB$}}
\newcommand{\bcalC}{\mbox{\boldmath $\calC$}}
\newcommand{\bcalD}{\mbox{\boldmath $\calD$}}
\newcommand{\bcalE}{\mbox{\boldmath $\calE$}}
\newcommand{\bcalF}{\mbox{\boldmath $\calF$}}
\newcommand{\bcalG}{\mbox{\boldmath $\calG$}}
\newcommand{\bcalH}{\mbox{\boldmath $\calH$}}
\newcommand{\bcalI}{\mbox{\boldmath $\calI$}}
\newcommand{\bcalJ}{\mbox{\boldmath $\calJ$}}
\newcommand{\bcalK}{\mbox{\boldmath $\calK$}}
\newcommand{\bcalL}{\mbox{\boldmath $\calL$}}
\newcommand{\bcalM}{\mbox{\boldmath $\calM$}}
\newcommand{\bcalN}{\mbox{\boldmath $\calN$}}
\newcommand{\bcalO}{\mbox{\boldmath $\calO$}}
\newcommand{\bcalP}{\mbox{\boldmath $\calP$}}
\newcommand{\bcalQ}{\mbox{\boldmath $\calQ$}}
\newcommand{\bcalR}{\mbox{\boldmath $\calR$}}
\newcommand{\bcalS}{\mbox{\boldmath $\calS$}}
\newcommand{\bcalT}{\mbox{\boldmath $\calT$}}
\newcommand{\bcalU}{\mbox{\boldmath $\calU$}}
\newcommand{\bcalV}{\mbox{\boldmath $\calV$}}
\newcommand{\bcalW}{\mbox{\boldmath $\calW$}}
\newcommand{\bcalX}{\mbox{\boldmath $\calX$}}
\newcommand{\bcalY}{\mbox{\boldmath $\calY$}}
\newcommand{\bcalZ}{\mbox{\boldmath $\calZ$}}

\newcommand{\sfA}{\mbox{$\mathsf A$}}
\newcommand{\sfB}{\mbox{$\mathsf B$}}
\newcommand{\sfC}{\mbox{$\mathsf C$}}
\newcommand{\sfD}{\mbox{$\mathsf D$}}
\newcommand{\sfE}{\mbox{$\mathsf E$}}
\newcommand{\sfF}{\mbox{$\mathsf F$}}
\newcommand{\sfG}{\mbox{$\mathsf G$}}
\newcommand{\sfH}{\mbox{$\mathsf H$}}
\newcommand{\sfI}{\mbox{$\mathsf I$}}
\newcommand{\sfJ}{\mbox{$\mathsf J$}}
\newcommand{\sfK}{\mbox{$\mathsf K$}}
\newcommand{\sfL}{\mbox{$\mathsf L$}}
\newcommand{\sfM}{\mbox{$\mathsf M$}}
\newcommand{\sfN}{\mbox{$\mathsf N$}}
\newcommand{\sfO}{\mbox{$\mathsf O$}}
\newcommand{\sfP}{\mbox{$\mathsf P$}}
\newcommand{\sfQ}{\mbox{$\mathsf Q$}}
\newcommand{\sfR}{\mbox{$\mathsf R$}}
\newcommand{\sfS}{\mbox{$\mathsf S$}}
\newcommand{\sfT}{\mbox{$\mathsf T$}}
\newcommand{\sfU}{\mbox{$\mathsf U$}}
\newcommand{\sfV}{\mbox{$\mathsf V$}}
\newcommand{\sfW}{\mbox{$\mathsf W$}}
\newcommand{\sfX}{\mbox{$\mathsf X$}}
\newcommand{\sfY}{\mbox{$\mathsf Y$}}
\newcommand{\sfZ}{\mbox{$\mathsf Z$}}

\newcommand{\balpha}{\mbox{\boldmath $\alpha$}}
\newcommand{\bbeta}{\mbox{\boldmath $\beta$}}
\newcommand{\bgamma}{\mbox{\boldmath $\gamma$}}
\newcommand{\bdelta}{\mbox{\boldmath $\delta$}}
\newcommand{\bepsilon}{\mbox{\boldmath $\epsilon$}}
\newcommand{\bvarepsilon}{\mbox{\boldmath $\varepsilon$}}
\newcommand{\bzeta}{\mbox{\boldmath $\zeta$}}
\newcommand{\boldeta}{\mbox{\boldmath $\eta$}}
\newcommand{\btheta}{\mbox{\boldmath $\theta$}}
\newcommand{\bvartheta}{\mbox{\boldmath $\vartheta$}}
\newcommand{\biota}{\mbox{\boldmath $\iota$}}
\newcommand{\bkappa}{\mbox{\boldmath $\kappa$}}
\newcommand{\blambda}{\mbox{\boldmath $\lambda$}}
\newcommand{\bmu}{\mbox{\boldmath $\mu$}}
\newcommand{\bnu}{\mbox{\boldmath $\nu$}}
\newcommand{\bxi}{\mbox{\boldmath $\xi$}}
\newcommand{\bpi}{\mbox{\boldmath $\pi$}}
\newcommand{\bvarpi}{\mbox{\boldmath $\varpi$}}
\newcommand{\brho}{\mbox{\boldmath $\rho$}}
\newcommand{\bvarrho}{\mbox{\boldmath $\varrho$}}
\newcommand{\bsigma}{\mbox{\boldmath $\sigma$}}
\newcommand{\bvarsigma}{\mbox{\boldmath $\varsigma$}}
\newcommand{\btau}{\mbox{\boldmath $\tau$}}
\newcommand{\bupsilon}{\mbox{\boldmath $\upsilon$}}
\newcommand{\bphi}{\mbox{\boldmath $\phi$}}
\newcommand{\bvarphi}{\mbox{\boldmath $\varphi$}}
\newcommand{\bchi}{\mbox{\boldmath $\chi$}}
\newcommand{\bpsi}{\mbox{\boldmath $\psi$}}
\newcommand{\bomega}{\mbox{\boldmath $\omega$}}

\newcommand{\bGamma}{\mbox{\boldmath $\Gamma$}}
\newcommand{\bDelta}{\mbox{\boldmath $\Delta$}}
\newcommand{\bTheta}{\mbox{\boldmath $\Theta$}}
\newcommand{\bLambda}{\mbox{\boldmath $\Lambda$}}
\newcommand{\bXi}{\mbox{\boldmath $\Xi$}}
\newcommand{\bPi}{\mbox{\boldmath $\Pi$}}
\newcommand{\bSigma}{\mbox{\boldmath $\Sigma$}}
\newcommand{\bUpsilon}{\mbox{\boldmath $\Upsilon$}}
\newcommand{\bPhi}{\mbox{\boldmath $\Phi$}}
\newcommand{\bPsi}{\mbox{\boldmath $\Psi$}}
\newcommand{\bOmega}{\mbox{\boldmath $\Omega$}}

\newcommand{\veca}{{\vec{\ba}}}
\newcommand{\vecb}{{\vec{\bb}}}
\newcommand{\vecc}{{\vec{\bc}}}
\newcommand{\vecd}{{\vec{\bd}}}
\newcommand{\vece}{{\vec{\bolde}}}
\newcommand{\vecf}{{\vec{\boldf}}}
\newcommand{\vecg}{{\vec{\bg}}}
\newcommand{\vech}{{\vec{\bh}}}
\newcommand{\veci}{{\vec{\bi}}}
\newcommand{\vecj}{{\vec{\bj}}}
\newcommand{\veck}{{\vec{\bk}}}
\newcommand{\vecl}{{\vec{\bl}}}
\newcommand{\vecm}{{\vec{\bm}}}
\newcommand{\vecn}{{\vec{\bn}}}
\newcommand{\veco}{{\vec{\bo}}}
\newcommand{\vecp}{{\vec{\bp}}}
\newcommand{\vecq}{{\vec{\bq}}}
\newcommand{\vecr}{{\vec{\br}}}
\newcommand{\vecs}{{\vec{\bs}}}
\newcommand{\vect}{{\vec{\bt}}}
\newcommand{\vecu}{{\vec{\bu}}}
\newcommand{\vecv}{{\vec{\bv}}}
\newcommand{\vecw}{{\vec{\bw}}}
\newcommand{\vecx}{{\vec{\bx}}}
\newcommand{\vecy}{{\vec{\by}}}
\newcommand{\vecz}{{\vec{\bz}}}

\newcommand{\vecxi}{{\vec{\bxi}}}
\newcommand{\vecphi}{{\vec{\bphi}}}
\newcommand{\vecvarphi}{{\vec{\bvarphi}}}
\newcommand{\vecbeta}{{\vec{\bbeta}}}
\newcommand{\vecdelta}{{\vec{\bdelta}}}
\newcommand{\vectheta}{{\vec{\btheta}}}

\newcommand{\Real}{\mathbb R}
\newcommand{\Complex}{\mathbb C}
\newcommand{\Natural}{\mathbb N}
\newcommand{\Integer}{\mathbb Z}


\newcommand{\bone}{\mbox{\boldmath $1$}}
\newcommand{\bzero}{\mbox{\boldmath $0$}}
\newcommand{\0}{{\bf 0}}

\newcommand{\be}{\begin{eqnarray}}
\newcommand{\ee}{\end{eqnarray}}
\newcommand{\bee}{\begin{eqnarray*}}
\newcommand{\eee}{\end{eqnarray*}}

\newcommand{\matrixb}{\left[ \begin{array}}
\newcommand{\matrixe}{\end{array} \right]}   

\newtheorem{theorem}{{Theorem}}
\newtheorem{lemma}{{Lemma}}
\newtheorem{corollary}{{Corollary}}
\newtheorem{definition}{{Definition}}
\newtheorem{proposition}{{Proposition}}
\newtheorem{remark}{{Remark}}
\newtheorem{example}{{Example}}

\newcommand{\argmax}{\operatornamewithlimits{\arg \max}}
\newcommand{\argmin}{\operatornamewithlimits{\arg \min}}

\newcommand{\mean}[1]{\left \langle #1 \right \rangle}
\newcommand{\ave}{\mathbb E}
\newcommand{\E}{\mathbb E}
\newcommand{\empha}[1]{{\color{red} \bf #1}}
\newcommand{\fracpartial}[2]{\frac{\partial #1}{\partial  #2}}
\newcommand{\incomplete}[1]{\textcolor{red}{#1}}

\def\doublespace{\renewcommand{\baselinestretch}{2}\large\normalsize}
\def\singlespace{\renewcommand{\baselinestretch}{1}\large\normalsize}
\def\onehalfspace{\renewcommand{\baselinestretch}{1.5}\large\normalsize}
\def\onequaterspace{\renewcommand{\baselinestretch}{1.3}\large\normalsize}
\def\threequaterspace{\renewcommand{\baselinestretch}{1.7}\large\normalsize}
\def\smallspace{\renewcommand{\baselinestretch}{-.9}\large\normalsize}
\def\tinyspace{\renewcommand{\baselinestretch}{-.7}\large\normalsize}

\newcommand{\tr} { \textrm{tr} }
\newcommand{\re} { \textrm{re} }
\newcommand{\diag} { \textrm{diag} }
\newcommand{\ddiag} { \textrm{ddiag} }
\newcommand{\off} { \textrm{off} }
\newcommand{\vectxt} { \textrm{vec} }

\newcommand{\lla}{\left\langle}
\newcommand{\rra}{\right\rangle}
\newcommand{\llbr}{\left\lbrack}
\newcommand{\rrbr}{\right\rbrack}
\newcommand{\llb}{\left\lbrace}
\newcommand{\rrb}{\right\rbrace}

\newcommand{\red}[1]{{\color{red}#1}}

\maketitle
\footnotetext[1]{These authors contributed equally.}
\footnotetext[2]{Corresponding author.}

\begin{abstract}
Recent advances in large pre-trained vision-language models have demonstrated remarkable performance on zero-shot downstream tasks. Building upon this, recent studies, such as CoOp and CoCoOp, have proposed the use of prompt learning, where context within a prompt is replaced with learnable vectors, leading to significant improvements over manually crafted prompts. However, the performance improvement for unseen classes is still marginal, and to tackle this problem, data augmentation has been frequently used in traditional zero-shot learning techniques. Through our experiments, we have identified important issues in CoOp and CoCoOp: the context learned through traditional image augmentation is biased toward seen classes, negatively impacting generalization to unseen classes. To address this problem, we propose adversarial token embedding to disentangle low-level visual augmentation features from high-level class information when inducing bias in learnable prompts. Through our novel mechanism called ``Adding Attributes to Prompt Learning", \texttt{AAPL}, we guide the learnable context to effectively extract text features by focusing on high-level features for unseen classes. We have conducted experiments across 11 datasets, and overall, \texttt{AAPL} shows favorable performances compared to the existing methods in few-shot learning, zero-shot learning, cross-dataset, and domain generalization tasks. The code is available at: \href{https://github.com/Gahyeonkim09/AAPL}{https://github.com/Gahyeonkim09/AAPL}
\end{abstract}      
\section{Introduction}

Recent research has shown significant improvements not only in model generalization performance through the use of large-scale vision-language models (VLMs), but also in zero-shot image classification performance~\cite{zhang2022contrastive,singh2022flava,yuan2021florence,yao2021filip,zhai2022lit}. It has been demonstrated that utilizing VLMs such as contrastive language-image pretraining (CLIP)~\cite{radford2021learning}, ALIGN~\cite{jia2021scaling}, Flamingo~\cite{alayrac2022flamingo}, etc., is effective in extracting image and text information for training classification models. The strengths of these VLMs have proven to be effective in prompt learning and handling both visual and textual information efficiently~\cite{shu2022test,lu2022prompt,zhou2022learning,zhou2022conditional}. 
CoOp~\cite{zhou2022learning} and CoCoOp~\cite{zhou2022conditional} have effectively produced learnable context vectors for classification weights via a text encoder (\eg, Transformer~\cite{vaswani2017attention}) along with CLIP. Specifically, CoCoOp~\cite{zhou2022conditional} has enabled the creation of class-specific classification weights by incorporating additional context information generated from images. In addition, visual prompt tuning (VPT)~\cite{jia2022vpt} demonstrated performance improvements in downstream tasks by introducing a small number of learnable parameters into the encoder layer of the Transformer along with image patches, without the need to replace or fine-tune the pre-trained transformer.

\begin{figure}[t]
  \centering
  \includegraphics[width=0.99\linewidth]{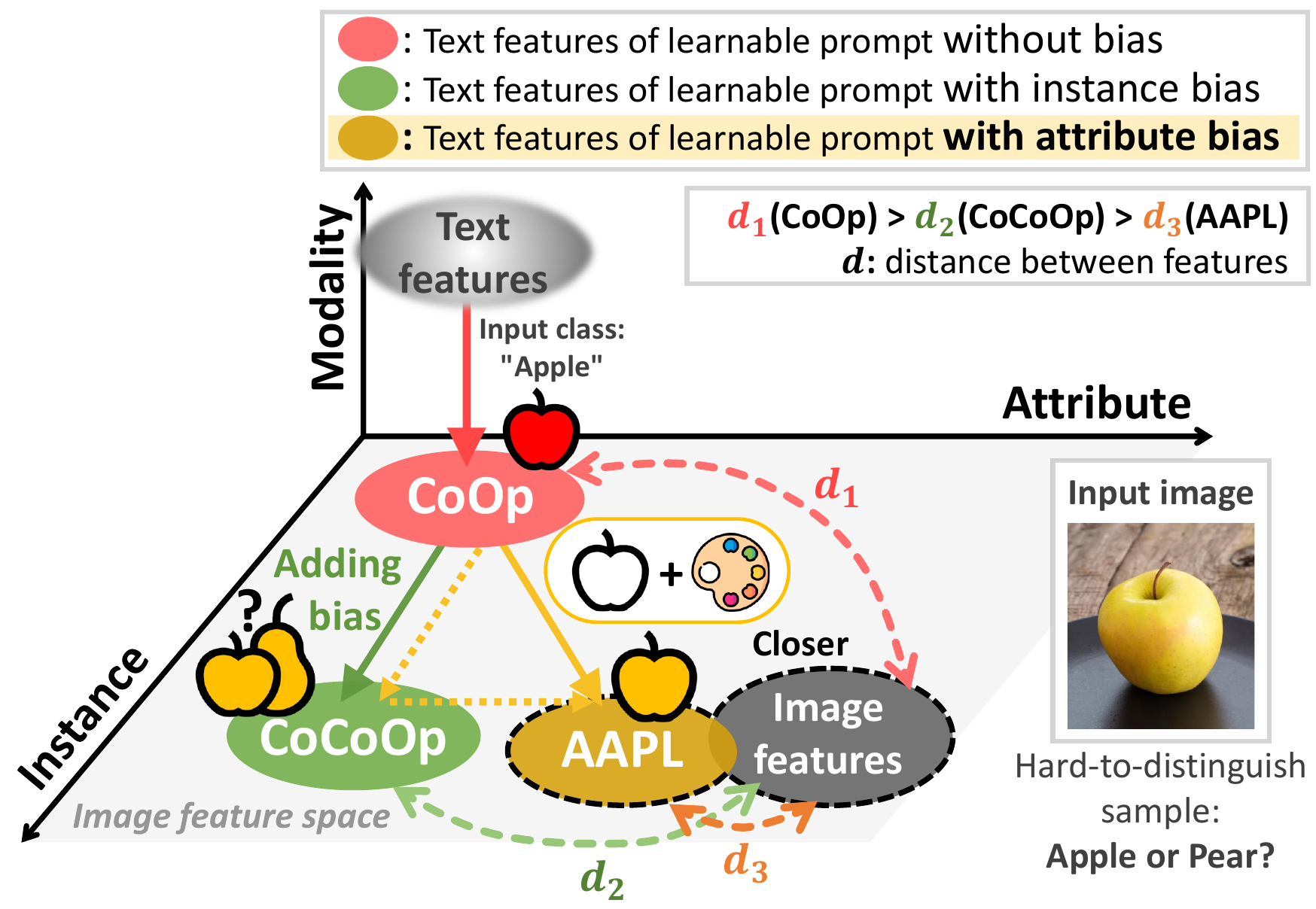}    
  \caption{\textbf{The illustration of \texttt{AAPL}.} 
    Training the learnable prompt on the class ``apple", since the training data mainly consists of red apples, leads to understanding apples as typically red. When a rare ``yellow apple” is input, the instance bias may overlook the yellow attribute and incorrectly predict it as a pear. However, \texttt{AAPL} extracts and decomposes attributes from the image, enhancing attribute-specific bias in the semantic features. 
    This enables robustly improved generalization performance across domains.
  } 
  \label{fig:fig1}
  \vspace{-8pt}  
\end{figure}

However, both CoOp~\cite{zhou2022learning} and VPT~\cite{jia2022vpt} have learnable parameters that are not manageable, especially in the case of CoCoOp~\cite{zhou2022conditional}, where it is unknown how the learnable vector will be shifted by the conditional bias based on particular information taken from the image that is added to the learnable context vector. This lack of management over learnable parameters can lead to unintentional bias in few-shot classification tasks or domain generalization tasks~\cite{ma2023understanding, khattak2023self, liu2023hierarchical}. To address this, we propose a new approach called \texttt{AAPL}, ``Adding Attributes to Prompt Learning”, as illustrated in Fig.~\ref{fig:fig1}.
In this context, augmentation generates a learnable bias that can be decomposed, with the augmented image serving as the visual prompt. 
Subsequent learning with visual-text prompts involves the use of a learnable context vector, which plays an adversarial role and mitigates unintended overfitting in downstream tasks~\cite{khattak2023self, liu2023hierarchical, ma2023understanding}. In summary, our contributions are as follows:
\begin{list}{$\diamond$}{\leftmargin=4mm \itemindent=0mm \itemsep=0mm}
  \item~We propose using augmented images as a visual prompt and introduce the concept of \textit{``delta meta token,''} which encapsulates attribute-specific information.
  \item~Employing \textit{delta meta token}, we conduct AdTriplet loss to make the conditional bias include the semantic feature of the class robustly, even in the presence of augmentation added to the learnable prompt through adversarial triplet loss.
  \item~We demonstrate performance improvements in base-to-new generalization tasks, cross-dataset tasks, and domain generalization tasks.
\end{list}

\label{sec:intro}

\section{Related Works}

\noindent\textbf{Vision-language models}~~~~Vision-language models (VLMs) using image-text pairs have shown superior capabilities over image-only models, especially in zero-shot transfer tasks for various downstream classification tasks~\cite{zhang2022contrastive,singh2022flava,yuan2021florence,yao2021filip,zhai2022lit}. Prominent models such as CLIP~\cite{radford2021learning} and ALIGN~\cite{jia2021scaling}, which have advanced through large-scale web data utilization, employ self-supervised learning for enhanced textual and visual alignment. In the embedding space, the contrastive loss draws matched image-text representation pairs closer, while it draws the representation of mismatched pairs farther away. Using this method, CLIP demonstrates exceptional zero-shot image recognition capabilities without the need for further fine-tuning. Our goal is to find efficient methods for applying pre-trained vision-language models to downstream applications, especially in prompt learning like CoOp~\cite{zhou2022learning} and CoCoOp~\cite{zhou2022conditional}.\\

\noindent\textbf{Prompt learning in vision-language models}~~~~The concept of prompt learning was initially proposed in the domain of natural language processing (NLP)~\cite{lester2021power,li2021prefix,liu2021p}. 
Unlike manually designing prompts, prompt learning research focuses on automatically selecting prompts during the fine-tuning stage.
Recently, this concept has been extended to the field of computer vision~\cite{zhou2022learning,lu2022prompt,khattak2023maple,zhu2023prompt,yao2023visual,wang2022learning,shu2022test,jia2022vpt}.
CoOp~\cite{zhou2022learning} introduced continuous prompt learning to the vision domain, applying pre-trained vision-language models to various tasks. Instead of using a manual prompt like ``a photo of a", they transformed the context word into a learnable context vector to optimize continuous prompts.
However, CoOp has limitations in generalizability due to overfitting on few-shot datasets. To address this, CoCoOp~\cite{zhou2022conditional} adds a conditional bias called \textit{meta token} extracted from image features to the learnable prompt. 
It shifts the focus from static to dynamic prompts, enabling optimization based on the characteristics of each instance rather than a specific class, consequently enhancing CoOp's domain generalization performance.
However, \textit{meta token} obtained from an image sample cannot be claimed to be completely robust against overfitting issues~\cite{ma2023understanding, khattak2023self, liu2023hierarchical}, and it is not interpretable because it is extracted from the shallow network, called \textit{metanet}, composed of Linear-ReLU-Linear layers. Therefore, we propose a new prompt learning method using image augmentation to leverage attribute-specific bias added to learnable prompts.\\

\noindent\textbf{Zero-shot learning}~~~~
Few-shot learning is the process of training on a small number of labeled samples before classifying the new images. In contrast, zero-shot learning (ZSL) aims to distinguish unseen classes by training exclusively on seen classes~\cite{chao2016empirical,xian2018zero}.
This is achieved by exclusively training on a set of base classes and utilizing side information, typically visual attributes like color, shape, and other features, shared with these unseen classes. This auxiliary information helps the machine understand language or concepts in a way humans do, enabling it to recognize unseen classes. 
The common methods~\cite{changpinyo2016synthesized,khan2023learning,mancini2021open,romera2015embarrassingly,xian2016latent} are learning the relation between a class embedding and the image feature, which represents this auxiliary information. 
However, these methods often exhibit a bias against unseen classes, known as ``seen-class bias"~\cite{xian2017zero}.
Other research efforts concentrate on enhancing visual-semantic embedding~\cite{cacheux2019modeling,zhang2017learning,jiang2019transferable}, or developing better image feature extractors~\cite{xu2020attribute,ji2018stacked}.
However, these methods usually assume a fixed set of auxiliary information, consisting of attributes labeled by humans. This assumption poses challenges, as labeling attributes is expensive, requires expert annotators, and is difficult to scale on large datasets.
Diverging from existing ZSL methods, our work focuses on adapting large vision-language models and employs techniques based on prompting. 

\label{sec:related}
\section{Methodology}
\label{sec:method}

\begin{figure*}[ht]
  \centering
  \includegraphics[width=0.80\textwidth]{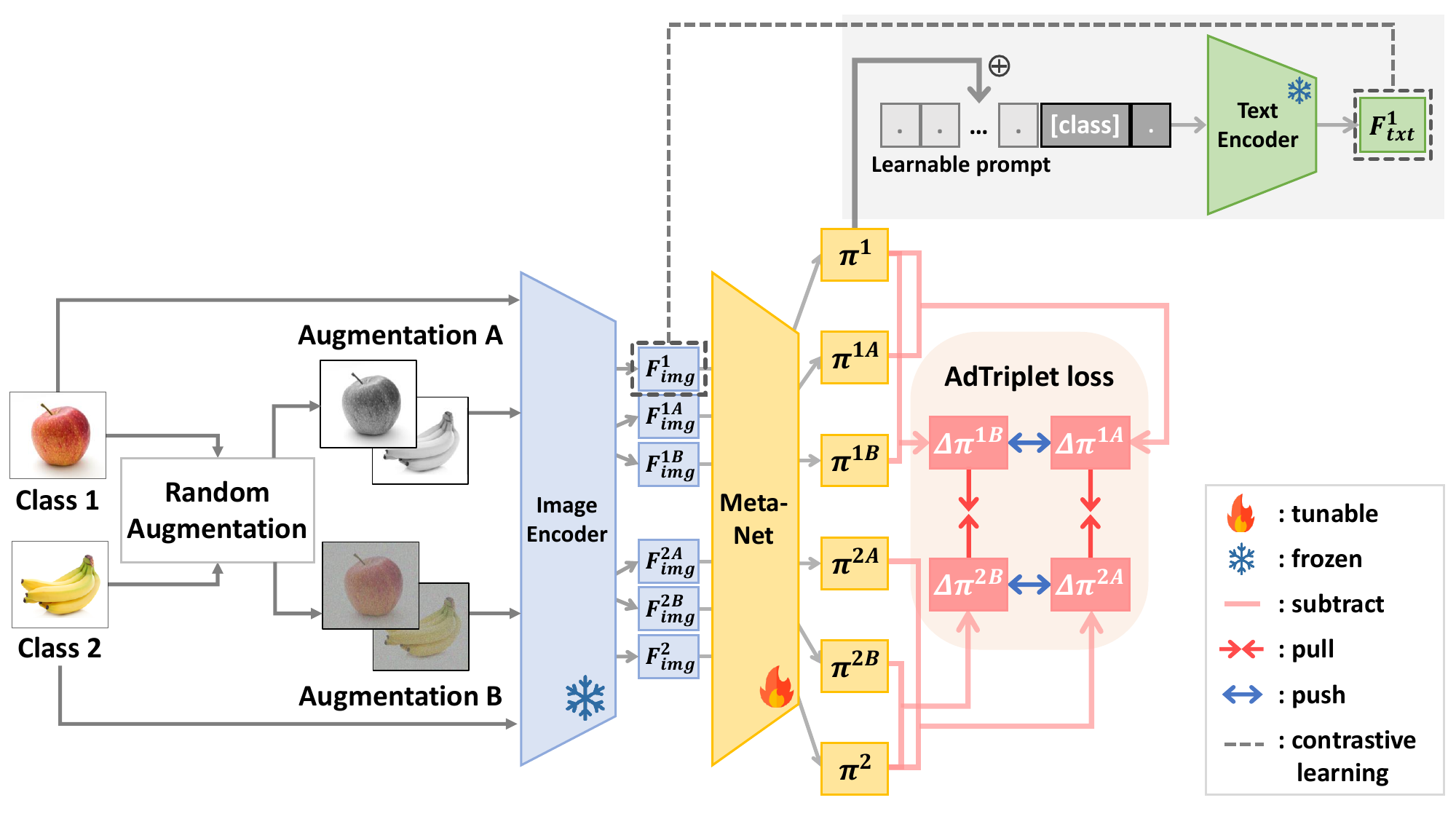}    
  \caption{\textbf{Overview of \texttt{AAPL}.} We apply two distinct random augmentations to the input images, each with the class labels 1 and 2. Once the image features are extracted from the pre-trained CLIP image encoder~\cite{radford2021learning}, they are passed through the \textit{metanet}~\cite{zhou2022conditional} to acquire the \textit{meta token}. These are then utilized to subtract the other meta tokens obtained from the augmented images for each class, resulting in \textit{delta meta tokens}. The goal is to instruct them to use these \textit{delta meta tokens} regardless of their classification. The \textit{delta meta tokens}, which are associated with the same augmentation, approach close within the embedding space using the AdTriplet loss, as shown in Eq.~\ref{eq:adtriplet_loss}. The \textit{delta meta tokens} acquire attribute-specific features, while the \textit{meta token} learns semantic features derived from image features, enabling the use of attribute-specific bias in the learnable prompt through the decomposed features.}
  \label{fig:overview_figure}
\end{figure*}

\subsection{Preliminaries}
\noindent\textbf{Prompt learning for CLIP}~~~~CLIP~\cite{radford2021learning} employs an image encoder based on ResNet~\cite{he2016deep} or ViT~\cite{dosovitskiy2020image} and a text encoder based on Transformer~\cite{vaswani2017attention} to extract features from images and text, respectively. These features are trained with a contrastive loss in the embedding space, aiming to maximize cosine similarities between paired modality features. When an input image $x$ is processed through the image encoder $f(\cdot)$, it generates an image feature $f(x)$. Using a prompt template like ``a photo of a \{{class}\}.", where the \{{class}\} token is substituted with the name of the $i$-th class, yields $K$ text features with corresponding weight vectors $\{{w_{i}}\}^{K}_{i=1}$ for the given $K$ class categories. The prediction probability for CoOp is as Eq.~\ref{eq:cocop_contrastive_loss}, where $sim(\cdot, \cdot)$ denotes cosine similarity and $\tau$ is a temperature parameter.
    \begin{equation}
        p(y|x) = \frac{\exp(sim(f(x), w_y)/\tau)}{\sum_{i=1}^{K} \exp(sim(f(x), w_i)/\tau)}
        \label{eq:cocop_contrastive_loss}
    \end{equation}    

\noindent\textbf{Conditional context optimization in prompt learning} ~~~~ CoOp~\cite{zhou2022learning} introduces context tokens as trainable vectors, $M$ learnable context, $\{{v_1, v_2,..., v_M}\}$, departing from a fixed template like ``a photo of a". The $i$-th class prompt, $t_i = \{{v_1, v_2,..., v_M, c_i}\}$, includes these vectors and word embeddings of the class name, $c_i$. Text features are generated from $t_i$ by CLIP text encoder $g(\cdot)$, which remained frozen throughout training. 
CoCoOp~\cite{zhou2022conditional} proposes instance-conditional context to prioritize individual input instances, reducing the overfitting of CoOp. This is done by using a \textit{metanet}, denoted as $h_{\theta}(\cdot)$ parameterized by $\theta$, to generate a conditional token for each input. 
Where $\pi = h_{\theta}(f(x))$ and $m \in \{{1, 2,..., M}\}$, each context token is obtained by $v_m(x) = v_m + \pi$. The prompt of the $i$-th class is conditioned on the input image feature, \ie, $t_i(x) = \{{v_1(x), v_2(x),..., v_M(x), c_i}\}$. Jointly updating context vectors $\{{v_m(x)}\}^{M}_{m=1}$ and \textit{metanet} during training ensures generalizability. The prediction probability for CoCoOp is as follows:
    \begin{equation}
        p(y|x) = \frac{\exp(\text{sim}(f(x), g(t_y(x)))/\tau)}{\sum_{i=1}^K \exp(\textit{sim}(f(x), g(t_i(x)))/\tau)}.
        \label{eq:cococop_contrastive_loss}
    \end{equation}
\label{subsection:preliminaries}

\subsection{Delta Meta Token}
\begin{figure}
    \centering
    \includegraphics[width=1.0\linewidth]{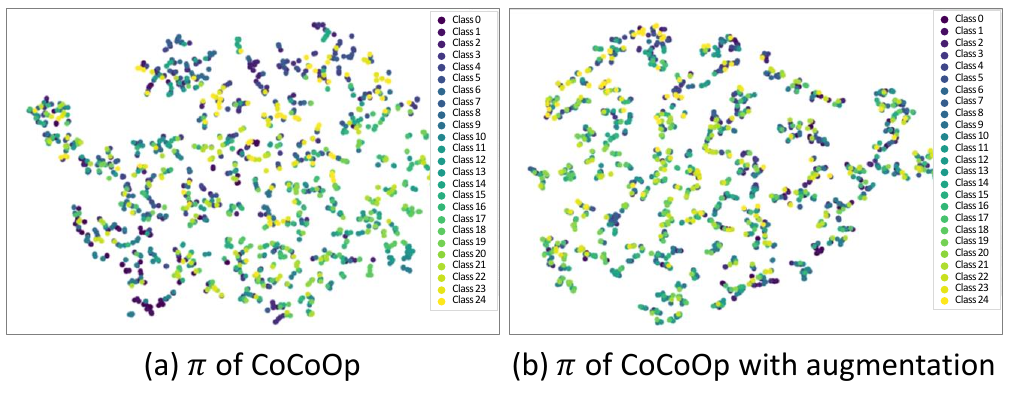}
    \caption{The comparison between \textit{meta tokens} of CoCoOp and \textit{meta tokens} of CoCoOp with random augmentation for FGVCAircraft dataset.} 
    \label{fig:aug_cocoop_figure}
\vspace{-1mm}
\end{figure}

\noindent\textbf{Effect of augmentation in CoCoOp}~~~~To investigate the effect of augmentation in prompt learning, we conducted a comparative experiment by adapting augmentation into CoCoOp~\cite{zhou2022conditional}. 
We added conditional bias from augmented images to the learnable prompt while maintaining other settings consistent with CoCoOp. 
As detailed in Table \ref{table:aug_cocoop}, incorporating augmentation leads to a decrease in base-to-new generalization accuracy compared to the original CoCoOp since the \textit{metanet} fails to extract the semantic features from the augmented images; thus extracting arbitrary noise rather than attribute-specific semantics. 
Additionally, as shown in Fig.~\ref{fig:aug_cocoop_figure}, it does not show a big difference in class clustering, indicating that the \textit{meta token} fails to capture the crucial semantic features for the classification. Consequently, this suggests that merely using augmentation in prompt learning might not enhance robustness or performance. It potentially leads to detrimental effects due to the \textit{metanet}'s inability to identify meaningful semantic features from the augmented images, focusing on instance-specific features rather than class semantics. To achieve optimal results, augmentation needs to be applied more carefully, ensuring that the conditional biases appropriately capture the semantic information of the class. \\
\label{subsection:delta_meta_token}                           

\definecolor{Gray2}{gray}{0.92}
\begin{table}
\centering
\footnotesize
\begin{tabular}{lccc}
\toprule[1pt]
Method & Base & New & HM \\
\midrule
CoOp~\cite{zhou2022learning} & \textbf{82.69} & 63.22 & 71.6 \\
CoCoOp~\cite{zhou2022conditional} & 80.47 & 71.69 & 75.83 \\
CoCoOp \footnotesize{with augmentation} & 79.25 & 70.89 & 74.38 \\
\rowcolor{Gray2} \textbf{\texttt{AAPL}} & 80.65 & \textbf{72.33} & \textbf{76.26} \\
\bottomrule[1pt]
\end{tabular}
\caption{The comparison of base-to-new generalization accuracy between \texttt{AAPL} and CoCoOp with augmentation. HM denotes harmonic mean score.}
\vspace{-3mm}
\label{table:aug_cocoop}
\end{table}             

\noindent\textbf{Delta meta token: detach attribute feature}~~~~CoCoOp~\cite{zhou2022conditional} improves the generalization performance of CoOp~\cite{zhou2022learning} by introducing \textit{metanet}, which outputs \textit{meta token} from image samples, then adds it to the learnable prompt. It focuses on learning about individual instance information rather than class information.  However, it's still unclear what information the \textit{meta token} contains, as the \textit{metanet} is a black box, and its shallow architecture leads to uncertain feature extraction. As shown in Fig.~\ref{fig:aug_cocoop_figure}, it fails to demonstrate clear clustering by neither augmentation type nor class. It shows that the \textit{meta token} does not effectively capture the semantic information of the class or the attribute of the input image sample. To address this issue and make it possible to add desired information to the learnable prompt, we propose the concept of a \textit{delta meta token}, the attribute-specific bias. The overview of \texttt{AAPL} is shown in Fig.~\ref{fig:overview_figure}.

To make a \textit{delta meta token}, two images of each of the two different classes are required, \eg, class~1 and class~2, as shown in Fig.~\ref{fig:overview_figure}. 
Two different augmentation types are randomly selected from 14 augmentations proposed in SimCLR~\cite{chen2020simple} for each pair of input images without any duplication, which is denoted as $Aug_A(\cdot)$ and $Aug_B(\cdot)$. 
Inspired by TextManiA~\cite{ye2023textmania}, which demonstrated the extraction of attribute information from text using Word Vector Analogy~\cite{ethayarajh2018towards, mikolov2013linguistic}, we generate \textit{delta meta token} by subtracting image features in the same class with different augmentation. \textit{Delta meta token} represents a difference vector from image features that contain augmentation information. They are generated at each iteration. The \textit{delta meta token} from an image $x$ of class 1 and $Aug_A(\cdot)$ can be written as follows:

\begin{equation}
    \Delta\pi^{1A} = h_{\theta}(f(Aug_A(x_1))) - h_{\theta}(f(x_1)) \label{eq:delta_meta_1A}.\\ 
\end{equation}

As TextManiA has shown, utilizing attributes containing semantic details derived from class information demonstrates its effectiveness in classification tasks. 
In other words, while the \textit{meta token} includes both class and attribute information, the \textit{delta meta token} preserves more specific image feature information associated with augmentation.
Adding decomposed auxiliary features to the learnable prompts, the \textit{delta meta token} can learn attribute information. We enable the learnable prompt to incorporate semantic features more abundantly, thus making the augmentation more effective. 
Similar to adversarial prompt learning for natural language processing~\cite{wu2022adversarial, nookala2023adversarial}, our method involves the adversarial interaction between class and attribute information, where the \textit{metanet} learns to extract attribute-related information from augmented image features. 
The more the learnable prompt learns the semantic feature information of the class, the better the classification performance.\\

\begin{figure*}
    \centering
    \includegraphics[width=1.0\linewidth]{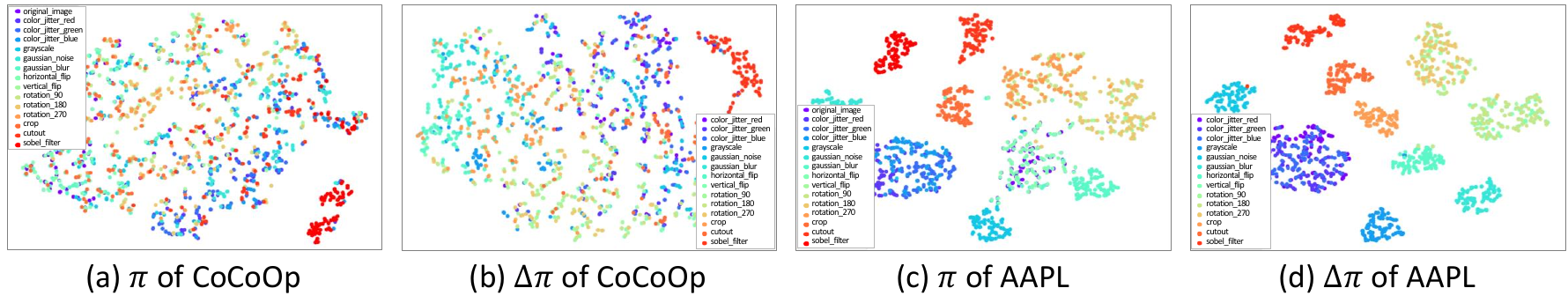}
    \caption{t-SNE visualization of \textit{meta token} and \textit{delta meta token} of CoCoOp~\cite{zhou2022conditional} and \texttt{AAPL} for FGVCAircraft dataset. The colors of the points represent the 14 different augmentations, and 100 data points from the validation set are used for this. $(a)$ and $(c)$ are the visualization of \textit{meta token}, $(b)$ and $(d)$ are the visualization of \textit{delta meta token}.}
    \label{fig:tsne_reuslts}
    \vspace{-1mm}
\end{figure*}

\begin{figure}
    \centering
    \includegraphics[width=0.8\linewidth]{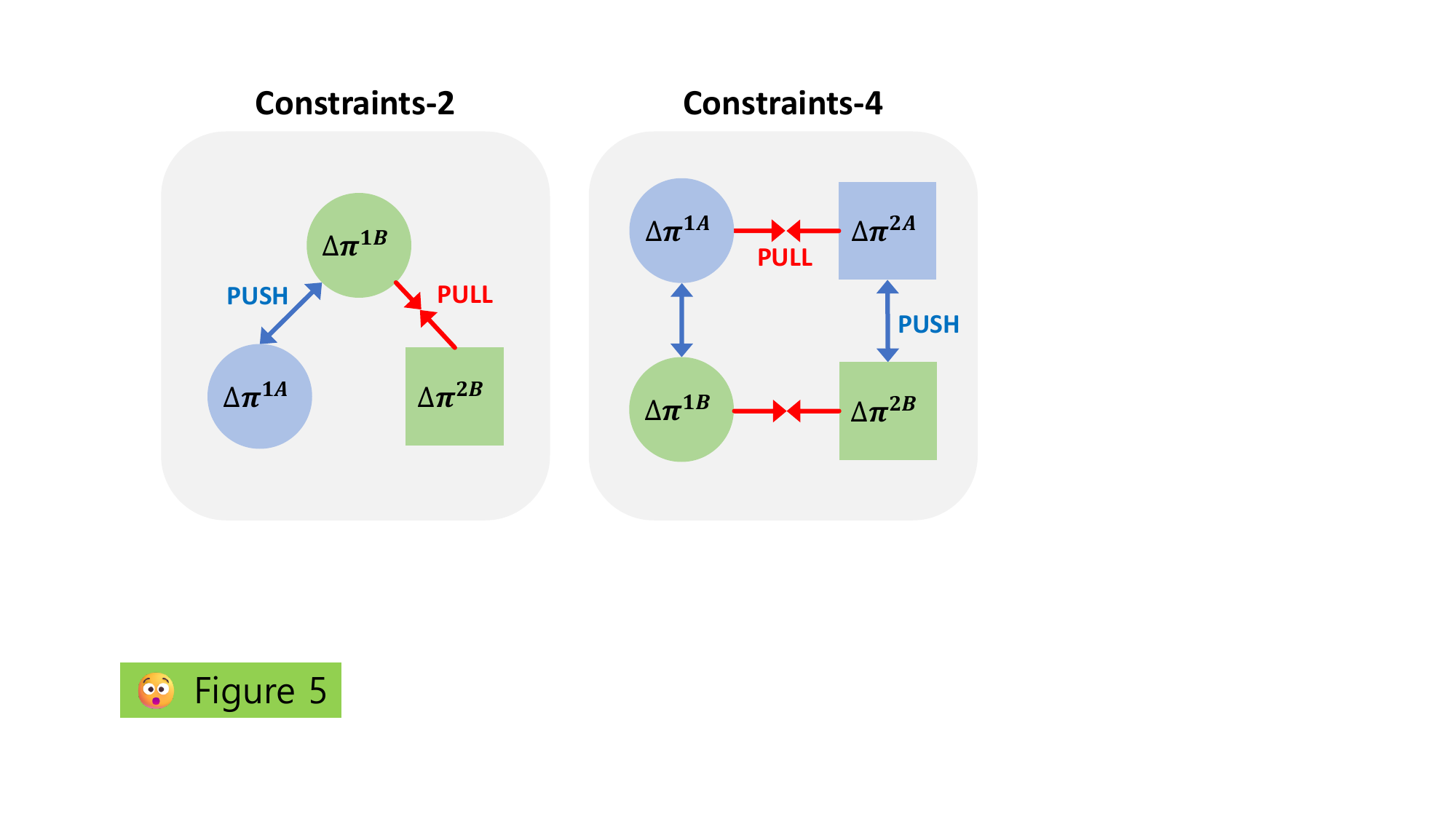}
    \caption{Comparison of the number of constraints of the AdTriplet loss. The constraints-2 setting's anchor is just one, \eg, $\Delta\pi^{1B}$, and the constraints-4 setting has two anchors, \eg, $\Delta\pi^{1A}$ and $\Delta\pi^{2B}$.}
    \label{fig:fig3}
    \vspace{-2mm}
\end{figure}

\noindent\textbf{Does the delta meta token have exact augmentation information?}~~~~In Fig.~\ref{fig:tsne_reuslts}, we used t-SNE to compare the validation results of \textit{metanet} of both CoCoOp~\cite{zhou2022conditional} and \texttt{AAPL}. It shows that CoCoOp fails to distinguish between augmentations compared to \texttt{AAPL}. As comparing Fig.~\ref{fig:tsne_reuslts} (c) and (d), while \textit{meta token} cannot perfectly discriminate 14 augmentations, \textit{delta meta token} shows almost perfect distinction, except for a few augmentations, \eg, vertical flip and rotations. This clustering result shows that the \textit{delta meta token} extracts more specific information about augmentation than the \textit{meta token}. As demonstrated in TextManiA~\cite{ye2023textmania}, for the textual case, subtraction between features can retain specific features. In the case of the image, we show that \textit{delta meta token} is more effective in making it contain the exact augmentation information. To the best of our knowledge, we are the first to employ feature decomposition through subtraction using visual features for prompt learning. It is noteworthy that, while the \textit{meta token} still retains information about the class, the \textit{delta meta token} accurately distinguishes between the semantic feature and the attribute feature.

\subsection{Adversarial Triplet Loss}
Using triplet loss~\cite{hoffer2015deep, sohn2016improved, weinberger2009distance, schroff2015facenet}, we can eliminate the remaining class-specific information in the \textit{delta meta token} while enhancing information related to augmentations. Training is conducted with 4 \textit{delta meta tokens}, \eg, $\Delta\pi^{1A}$, $\Delta\pi^{1B}$, $\Delta\pi^{2A}$, and $\Delta\pi^{2B}$, in the embedding space, aiming to increase the distance between vectors of the same class while minimizing it for the same augmentation. 
For instance, considering anchor as $\Delta\pi^{1A}$, its positive pair is $\Delta\pi^{2A}$, which has a different class but the same augmentation. In contrast, $\Delta\pi^{1B}$ is considered a negative pair because it has the same class but a different augmentation.
The distance between the anchor and the negative pair should be greater than the distance between the anchor and the positive pair. The Euclidean distance is denoted as $\|\ {\cdot}\|_2$, and the margin of the triplet loss is denoted as $m$ in Eq.~\ref{eq:triplet_loss1}.

\begin{multline}
L_{triplet}(x, x^{+}, x^{-} ;\Delta\pi^{1A}, \Delta\pi^{2A},\Delta\pi^{1B}) \\
  = max (0,\ {\|x - x^{+} \|_2 - \|x - x^{-}\|_2} + m ) \\
  = max (0,\ {\|\Delta\pi^{1A} - \Delta\pi^{2A}\|_2 - \|\Delta\pi^{1A} - \Delta\pi^{1B}\|_2} + m )      
  \label{eq:triplet_loss1}
\end{multline}

Thus, we introduce the Adtriplet loss, which adversarially trains the model to prioritize the alignment of augmentation information over class information. This loss is updated alongside the classification loss, specifically the cross-entropy loss.
The AdTriplet loss is used as constraints-4, as illustrated in Fig.~\ref{fig:fig3}, to make the connection between the class information domain and augmentation attribute domain more balanced~\cite{kim2018co}.

\begin{multline}
    L_{AdTriplet} = L_{triplet}^1(\Delta\pi^{1A}, \Delta\pi^{2A},\Delta\pi^{1B}) \\+ L_{triplet}^2(\Delta\pi^{2B}, \Delta\pi^{1B}, \Delta\pi^{2A}) 
    \label{eq:adtriplet_loss} 
\end{multline} 

Cross-entropy loss is computed following the same method as CoCoOp~\cite{zhou2022conditional}. 
To ensure fairness between the training and test phases, only one input image label is used for cross-entropy loss calculation.
The final training loss function is as follows: 
\begin{equation}
    L_{total} = \alpha * L_{AdTriplet} + \beta * L_{CE},
\label{eq:total_loss}
\end{equation} 
where $\alpha$ and $\beta$ are hyper-parameters for scaling. 
In Sec.~\ref{sec:experiment}, we provide detailed information on parameter tuning.

\label{subsection:adversarial_triplet_loss}

\section{Experiments}
\label{sec:experiment}
\definecolor{Gray}{gray}{0.92}

\subsection{Experimental Settings}
\noindent\textbf{Datasets}~~We use 11 classification datasets based on CLIP~\cite{radford2021learning}, CoOp~\cite{zhou2022learning}, and CoCoOp~\cite{zhou2022conditional} for base-to-new generalization and cross-dataset transfer: ImageNet~\cite{deng2009imagenet} and Caltech101~\cite{fei2004learning} for generic object classification, OxfordPets~\cite{parkhi2012cats}, StanfordCars~\cite{krause20133d}, Flowers102~\cite{nilsback2008automated}, Food101~\cite{bossard2014food} and FGVCAircraft~\cite{maji2013fine} for fine-grained image recognition, EuroSAT~\cite{helber2019eurosat} for satellite image classification, UCF101~\cite{soomro2012ucf101} for action classification, DTD~\cite{cimpoi2014describing} for texture classification, and SUN397~\cite{xiao2010sun} for scene recognition. 
For domain generalization experiments, we use ImageNet~\cite{deng2009imagenet} as the source dataset and 4 other ImageNet-based datasets, \ie, ImageNetV2~\cite{recht2019imagenet}, ImageNetSketch~\cite{wang2019learning}, ImageNet-A~\cite{hendrycks2021natural}, and ImageNet-R~\cite{hendrycks2021many}, as the target datasets, which each contain a different kind of domain shift. 

\noindent\textbf{Baselines}~~We compare \texttt{AAPL} with 3 baseline methods: the zero-shot CLIP~\cite{radford2021learning}, CoOp~\cite{zhou2022learning}, and CoCoOp~\cite{zhou2022conditional}. 
CLIP uses the hand-crafted template ``a photo of a \{class\}" to generate the prompts for knowledge transfer.
CoOp learns a static prompt that replaces the hand-crafted prompts with the learnable vectors.
CoCoOp generates dynamic prompts by adding the image-conditional prompts to the learnable prompts in CoOp.

\noindent\textbf{Training details}~~Our implementation is based on CoCoOp~\cite{zhou2022conditional}. We employ the pre-trained ViT-B/16 model from CLIP~\cite{radford2021learning} as the backbone. We fix the context length to 4 and initialize the context vectors randomly. The presented results are the mean values obtained from experiments conducted with three random seeds. We follow the training epochs, batch sizes, and schedules as prescribed by CoCoOp. In the context of few-shot learning, we confine evaluation to the maximum shot, \ie, 16 shots, considered by CoOp. For evaluation, we use the model from the last epoch. The parameter size of \texttt{AAPL} is the same as CoCoOp, and the hyper-parameter $m$ in Eq.~\ref{eq:triplet_loss1} is set to 0.2.
\label{subsection: Experimental Setting}

\begin{table}[h]
\centering
\small
\begin{adjustbox}{width=0.99\linewidth}
\begin{tabular}{lc|ccccc}
\toprule[1pt]
\multirow{2}{15mm}{Dataset} & & CLIP & CoOp & CoCoOp & \textbf{\texttt{AAPL}} & \multirow{2}*{$\Delta$} \\
 & & \cite{radford2021learning} & \cite{zhou2022learning} & \cite{zhou2022conditional} & (Ours) & \\
\midrule
\multirow{3}{15mm}{Average on 11 datasets}& Base & 69.34 & \textbf{82.69} & 80.47 & 80.27 & \textcolor{BrickRed}{-0.20} \\
                                          & Novel & \textbf{74.22} & 63.22 & 71.69 & 72.17 & \textcolor{NavyBlue}{+0.48} \\
                                          & HM & 71.70 & 71.66 & 75.83 & \textbf{76.01} & \textcolor{NavyBlue}{+0.18} \\ 
\midrule
\multirow{3}{15mm}{ImageNet}& Base & 72.43 &  76.47 & 75.98 &\textbf{ 76.53} & \textcolor{NavyBlue}{+0.55} \\
                            & Novel & 68.14 & 67.88 & 70.43 & \textbf{70.57} & \textcolor{NavyBlue}{+0.14} \\
                            & HM & 70.22 & 71.92 & 73.10 & \textbf{73.43} & \textcolor{NavyBlue}{+0.33} \\
\midrule
\multirow{3}{15mm}{Caltech101}& Base & 96.84 & \textbf{98.00} & 97.96 & 97.87 & \textcolor{BrickRed}{-0.09} \\
                            & Novel & 94.00 & 89.81 & 93.81 & \textbf{95.10} & \textcolor{NavyBlue}{+1.29} \\
                            & HM & 95.40 & 93.73 & 95.84 & \textbf{96.46} & \textcolor{NavyBlue}{+0.62} \\
\midrule
\multirow{3}{15mm}{OxfordPets}& Base & 91.17 & 93.67 & 95.20 & \textbf{95.63} & \textcolor{NavyBlue}{+0.43} \\
                            & Novel & 97.26 & 95.29 & \textbf{97.69} & 97.40 & \textcolor{BrickRed}{-0.29} \\
                            & HM & 94.12 & 94.47 & 96.43 & \textbf{96.51} & \textcolor{NavyBlue}{+0.08} \\
\midrule
\multirow{3}{15mm}{Stanford Cars}& Base & 63.37 & \textbf{78.12} & 70.49 & 70.33 & \textcolor{BrickRed}{-0.16} \\
                            & Novel & \textbf{74.89} & 60.40 & 73.59 & 73.50 & \textcolor{BrickRed}{-0.09} \\
                            & HM & 68.65 & 68.13 & \textbf{72.01} & 71.88 & \textcolor{BrickRed}{-0.13} \\
\midrule
\multirow{3}{15mm}{Flowers102}& Base & 72.08 &\textbf{ 97.60} & 94.87 & 95.10 & \textcolor{NavyBlue}{+0.23} \\
                            & Novel & \textbf{77.80} & 59.67 & 71.75 & 70.63 & \textcolor{BrickRed}{-1.12} \\
                            & HM & 74.83 & 74.06 & \textbf{81.71} & 81.06 & \textcolor{BrickRed}{-0.65} \\
\midrule
\multirow{3}{15mm}{Food101}& Base & 90.10 & 88.33 & \textbf{90.70} & \textbf{90.70} & \textcolor{NavyBlue}{+0.00} \\
                            & Novel & 91.22 & 82.26 & 91.29 & \textbf{91.60} & \textcolor{NavyBlue}{+0.31} \\
                            & HM & 90.66 & 85.19 & 90.99 & \textbf{91.15} & \textcolor{NavyBlue}{+0.16} \\
\midrule
\multirow{3}{15mm}{FGVC Aircraft}& Base & 27.19 & \textbf{40.44} & 33.41 & 34.07 & \textcolor{NavyBlue}{+0.66} \\
                            & Novel & \textbf{36.29} & 22.30 & 23.71 & 24.17 & \textcolor{NavyBlue}{+0.46} \\
                            & HM & \textbf{31.09} & 28.75 & 27.74 & 28.28 & \textcolor{NavyBlue}{+0.54} \\
\midrule
\multirow{3}{15mm}{SUN397}& Base & 69.36 & 80.60 & \textbf{79.74} & 79.65 & \textcolor{BrickRed}{-0.09} \\
                            & Novel & 75.35 & 65.89 & 76.86 & \textbf{76.90} & \textcolor{NavyBlue}{+0.04} \\
                            & HM & 72.23 & 72.51 & \textbf{78.27} & 78.25 & \textcolor{BrickRed}{-0.02} \\
\midrule
\multirow{3}{15mm}{DTD}& Base & 53.24 & \textbf{79.44} & 77.01 & 73.90 & \textcolor{BrickRed}{-3.11} \\
                            & Novel & \textbf{59.90} & 41.18 & 56.00 & 53.43 & \textcolor{BrickRed}{-2.57} \\
                            & HM & 56.37 & 54.24 & \textbf{64.85} & 62.02 & \textcolor{BrickRed}{-2.83} \\
\midrule
\multirow{3}{15mm}{EuroSAT}& Base & 56.48 & \textbf{92.19} & 87.49 & 87.00 & \textcolor{BrickRed}{-0.49} \\
                            & Novel & 64.05 & 54.74 & 60.04 & \textbf{66.30} & \textcolor{NavyBlue}{+6.26} \\
                            & HM & 60.03 & 68.69 & 71.21 & \textbf{75.25} & \textcolor{NavyBlue}{+4.04} \\
\midrule
\multirow{3}{15mm}{UCF101}& Base & 70.53 & \textbf{84.69} & 82.33 & 82.20 & \textcolor{BrickRed}{-0.13} \\
                            & Novel & \textbf{77.50} & 56.05 & 73.45 & 74.27 & \textcolor{NavyBlue}{+0.82} \\
                            & HM & 73.85 & 67.46 & 77.64 & \textbf{78.03} & \textcolor{NavyBlue}{+0.39} \\
\bottomrule[1pt]
\end{tabular}
\end{adjustbox}
\caption{\textbf{Base-to-new generalization experiment compared to baselines.} 
The model is trained from the base classes (16 shots) and evaluated in new classes. HM denotes the harmonic mean. $\Delta$ is the difference between \texttt{AAPL} and CoCoOp. The \textbf{bold} highlighting indicates the highest performance scores.}
\label{table:basetonew}
\end{table}

\subsection{Generalization from Base-to-New Classes}
We divided the classes equally into two groups, one for the base classes and another for the new classes, \ie, unseen classes, just like in CoCoOp~\cite{zhou2022conditional}. Learning-based models are trained solely on base classes. In few-shot learning, the model is evaluated with the base classes, whereas in zero-shot learning, it is evaluated with the new classes to test the model's generalizability.
In this task, we set hyper-parameters $\alpha$ and $\beta$ to 0.2 and 1. Table~\ref{table:basetonew} presents the performance results of \texttt{AAPL} compared to the baseline.
\texttt{AAPL} outperformed in 7 out of 11 datasets, with the harmonic mean of total dataset accuracy exceeding that of CoCoOp. However, performance on the DTD~\cite{cimpoi2014describing} was significantly lower. The geometrical augmentations, especially flips and rotations, appear to have minimal effect on \texttt{AAPL}, as they do not significantly alter the appearance of the original images in the context of texture. This demonstrates that the effectiveness of \texttt{AAPL} varies across different datasets. 
\label{subsection: Generalization from Base-to-New Classes}

\subsection{Cross-Dataset Transfer}
To assess the robustness and adaptability of \texttt{AAPL}, we tested its generalization ability across datasets by training it on all 1000 ImageNet classes and then applying it on the other 10 datasets, as shown in Table~\ref{table:cross_dataset}. 
We assume that the model can learn semantic information about image features by learning precise attributes. 
To evaluate this, we increased the model's focus on learning augmentation information by setting both hyper-parameters, $\alpha$ and $\beta$, to 1 in this experiment and afterward. \texttt{AAPL} achieves higher generalization in 3 datasets: OxfordPets~\cite{parkhi2012cats}, FGVCAircraft~\cite{maji2013fine}, and UCF101~\cite{soomro2012ucf101}, compared to CoCoOp~\cite{zhou2022conditional}. However, the performance on DTD~\cite{cimpoi2014describing} and EuroSAT~\cite{helber2019eurosat} was noticeably poorer than other datasets. This suggests that these datasets are vulnerable to \texttt{AAPL}'s augmentation-based prompt learning. 
These datasets are not object-centric but rather possess global features, \eg, long-distance satellite images and texture images. Extracting specific attributes from these datasets is challenging due to their unique characteristics. 
\label{subsection: Cross-Dataset Transfer}

\begin{table}[h]
\centering
\begin{adjustbox}{width=1.0\linewidth}
\begin{tabular}{l@{\,}c@{\hspace{6pt}}c@{\hspace{6pt}}c@{\hspace{6pt}}c@{\hspace{6pt}}c@{\hspace{6pt}}c@{\hspace{6pt}}c
                @{\hspace{6pt}}c@{\hspace{6pt}}c@{\hspace{6pt}}c@{\hspace{6pt}}c@{\,\hspace{6pt}}c}
\toprule[1pt]
& Source & \multicolumn{11}{c}{Target} \\ 
\cmidrule(l{2pt}r{2pt}){2-0} \cmidrule(l{2pt}r{2pt}){3-12} 
& \rotatebox{90}{ImageNet} & \rotatebox{90}{Caltech101} & \rotatebox{90}{OxfordPets} & \rotatebox{90}{StanfordCars} & \rotatebox{90}{Flowers102} & \rotatebox{90}{Food101} & \rotatebox{90}{FGVCAircraft} & \rotatebox{90}{SUN397} & \rotatebox{90}{DTD} & \rotatebox{90}{EuroSAT} & \rotatebox{90}{UCF101} & \textit{\rotatebox{90}{Average}} \\
\midrule
        CoOp & \textbf{71.51}& 93.70 & 89.14 & 64.51 & 68.71 & 85.30 & 18.47 & 64.15 & 41.92 & 46.39 & 66.55 & 63.88 \\
        CoCoOp & 71.02 &\textbf{ 94.43} & 90.14 & \textbf{65.32} & \textbf{71.88} & \textbf{86.02} & 22.94 & \textbf{67.36} & \textbf{45.73} & \textbf{45.37} & 68.21 & \textbf{65.74} \\
\rowcolor{Gray} \textbf{\texttt{AAPL}} & 71.37 & 94.17 & \textbf{90.73} & 65.10 & 71.67 & 86.00 & \textbf{23.03} & 66.80 & 44.80 & 41.83 & \textbf{69.30} & 65.34 \\

\bottomrule[1pt]
\end{tabular}
\end{adjustbox}
\caption{\textbf{Cross-dataset transfer experiment.} The model is trained on the entire class of ImageNet (16 shots) and evaluated on the other 10 datasets.}
\vspace{-1mm}
\label{table:cross_dataset} 
\end{table}

\subsection{Domain Generalization}
For domain generalization, we trained our model on the whole ImageNet dataset, same as in Sec.~\ref{subsection: Cross-Dataset Transfer}, and evaluated it on 4 datasets that represent a domain shift from ImageNet (\eg, ImageNetV2~\cite{recht2019imagenet}, ImageNetSketch~\cite{wang2019learning}, ImageNet-A~\cite{hendrycks2021natural}, and ImageNet-R~\cite{hendrycks2021many}). The comparison of these tests is presented in Table~\ref{table:domain_generalization}. We achieved better performance on all datasets except for ImageNet-A. This demonstrates that attribute-specific bias effectively deals with domain shift. \\
\label{subsection: Domain Generalization}

\begin{table}[h]
\centering
\begin{adjustbox}{width=0.95\linewidth}
\begin{tabular}{lcccccc}
\toprule[1pt]
& Source & \multicolumn{4}{c}{Target} \\ 
\cmidrule(l{2pt}r{2pt}){2-0} \cmidrule(l{2pt}r{2pt}){3-6} 
& ImageNet & -V2 & -S & -A & -R & \textit{Avg.} \\
\midrule
CLIP & 66.73 & 60.83 & 46.15 & 47.77 & 73.96 & 57.18 \\
CoOp & \textbf{71.51} & \textbf{64.20 }& 47.99 & 49.71 & 75.21 & 59.28 \\
CoCoOp & 71.02 & 64.07 & 48.75 & \textbf{50.63} & 76.18 & 59.91  \\
\rowcolor{Gray} \textbf{\texttt{AAPL}} & 71.37 & \textbf{64.20} & \textbf{48.80} & 50.60 & \textbf{76.87} & \textbf{60.12} \\
\bottomrule[1pt]
\end{tabular}
\end{adjustbox}
\caption{\textbf{Domain Generalization experiment.} The model is trained on the entire class of ImageNet (16 shots) and evaluated on four different ImageNet-based datasets, including domain shifts. }
\label{table:domain_generalization} 
\end{table}

\subsection{Augmentation Profiling}
\begin{figure}[ht]
    \centering
    \includegraphics[width=0.95\linewidth]{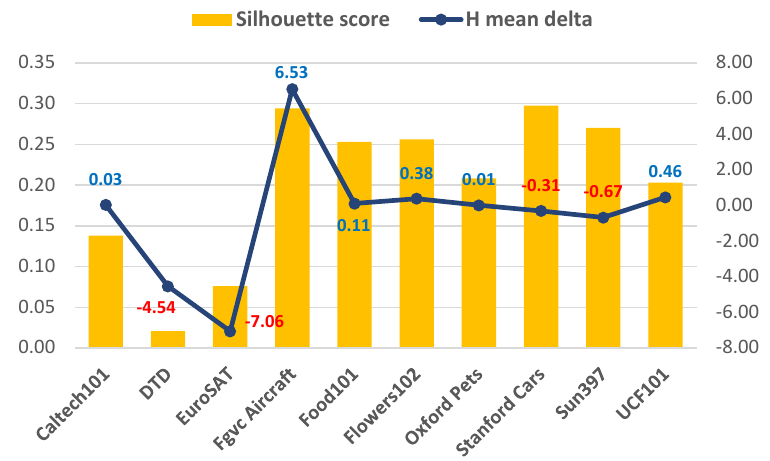}
    \caption{\textbf{The correlation between silhouette score and generalization performance.}
    Silhouette score and the difference in harmonic mean accuracy for zero-shot classification between CoCoOp and \texttt{AAPL}}
    \label{fig:fig_sil_score}
    \vspace{-2pt}
\end{figure}

\noindent\textbf{Why should the delta meta token learn about attributes rather than class information?}~~~~
To assess the effectiveness of learning attributes, we compared the silhouette scores~\cite{shahapure2020cluster} based on augmentation types. 
The silhouette score evaluates how well data points are clustered, considering both cohesion (proximity within the same cluster) and separation (distance from the nearest neighboring cluster).
The silhouette score $S(i)$ for data point $i$, is calculated as follows: $S(i) = \frac{{b(i)-a(i)}}{max\{{a(i), b(i)}\}}$, where $a(i)$ is the average distance of $i$ to all other data points in the same cluster, and $b(i)$ is the average distance of $i$ to the data points in the nearest cluster that $i$ does not belong to.
A higher silhouette score indicates better clustering. In other words, datasets that effectively learn information about augmentations from the Adtriplet loss have higher silhouette scores. 
As shown in Fig.~\ref{fig:fig_sil_score}, the zero-shot classification performance of \texttt{AAPL} generally improves. However, there is a sharp decrease in performance for DTD~\cite{cimpoi2014describing}  and EuroSAT~\cite{helber2019eurosat}. 
This suggests that datasets that cannot effectively extract augmentation information do not perform well. 
Training precise attributes to \textit{delta meta token} is crucial for zero-shot classification, and it's evident that determining what information to add to the learnable prompt is highly important for datasets sensitive to \texttt{AAPL}.\\

\noindent\textbf{Which dataset is vulnerable for \texttt{\textbf{AAPL}}?}~~~~
To assess the impact of various datasets on the evaluation of learning attribute features, we applied \texttt{AAPL}'s proposed AdTriplet loss and the traditional triplet loss method. 
Unlike the AdTriplet loss, the traditional triplet loss trains the \textit{delta meta token} to cluster classes rather than augmentation types. 
As shown in Table~\ref{table:triplet_vs_adtriplet}, when utilizing Adtriplet loss across 6 datasets, performance improvement was observed compared to using triplet loss. Particularly, FGVCAircraft~\cite{maji2013fine} exhibited approximately a 7\% higher performance improvement with the triplet loss. Utilizing the traditional triplet loss method means that the \textit{delta meta token} is trained to bring the same class together regardless of augmentation type. Consequently, datasets that showed improved performance on AdTriplet loss have a higher dependency on class information. 
Triplet loss-trained \textit{delta meta token} extracts class-related information, causing \textit{meta token} to contribute noisy features rather than class semantic features when added to prompts. In contrast, AdTriplet loss-trained tokens focus on extracting the class semantic features. Datasets with AdTriplet loss perform well because they rely more on the class information. This highlights the advantage of \texttt{AAPL} based on the dataset's characteristics. \\

\begin{table}
\centering
\small
\begin{adjustbox}{width=1.0\linewidth}
\begin{tabular}{l@{\quad}c@{\,\hspace{6pt}}c@{\,\hspace{6pt}}c@{\,\hspace{6pt}}c@{\,\hspace{6pt}}c@{\,\hspace{6pt}}c@{\,\hspace{6pt}}c
                @{\,\hspace{6pt}}c@{\,\hspace{6pt}}c@{\,\hspace{6pt}}c@{\,\hspace{6pt}}c@{\,\hspace{6pt}}c}
\toprule[1pt]
& \rotatebox{90}{ImageNet} & \rotatebox{90}{Caltech101} & \rotatebox{90}{OxfordPets} & \rotatebox{90}{StanfordCars} & \rotatebox{90}{Flowers102} & \rotatebox{90}{Food101} & \rotatebox{90}{FGVCAircraft} & \rotatebox{90}{SUN397} & \rotatebox{90}{DTD} & \rotatebox{90}{EuroSAT} & \rotatebox{90}{UCF101} & \rotatebox{90}{\textit{Average}}\\
\midrule
        Triplet & \textbf{73.44}& 95.81 & 96.18 & \textbf{72.22} & 80.65 & 90.70 & 27.97 & \textbf{78.34} & \textbf{61.73} & 64.15 & \textbf{78.78} & 74.54 \\
        \rowcolor[gray]{0.92}AdTriplet & 73.09 &\textbf{96.87} &\textbf{96.44} & 71.70 & \textbf{82.09} & \textbf{91.10} & \textbf{34.27} & 77.60 & 60.31 & \textbf{65.16} & 78.10 & \textbf{75.16} \\
\bottomrule[1pt]
\end{tabular}
\end{adjustbox}
\caption{\textbf{\texttt{AAPL} with Triplet and AdTriplet loss.} The comparison of harmonic means of base-to-new generalization accuracy between \texttt{AAPL} trained with AdTriplet loss and traditional Triplet loss.}
\vspace{-10pt}
\label{table:triplet_vs_adtriplet} 
\end{table}

\begin{figure}[t]
    \centering
    \includegraphics[width=0.99\linewidth]{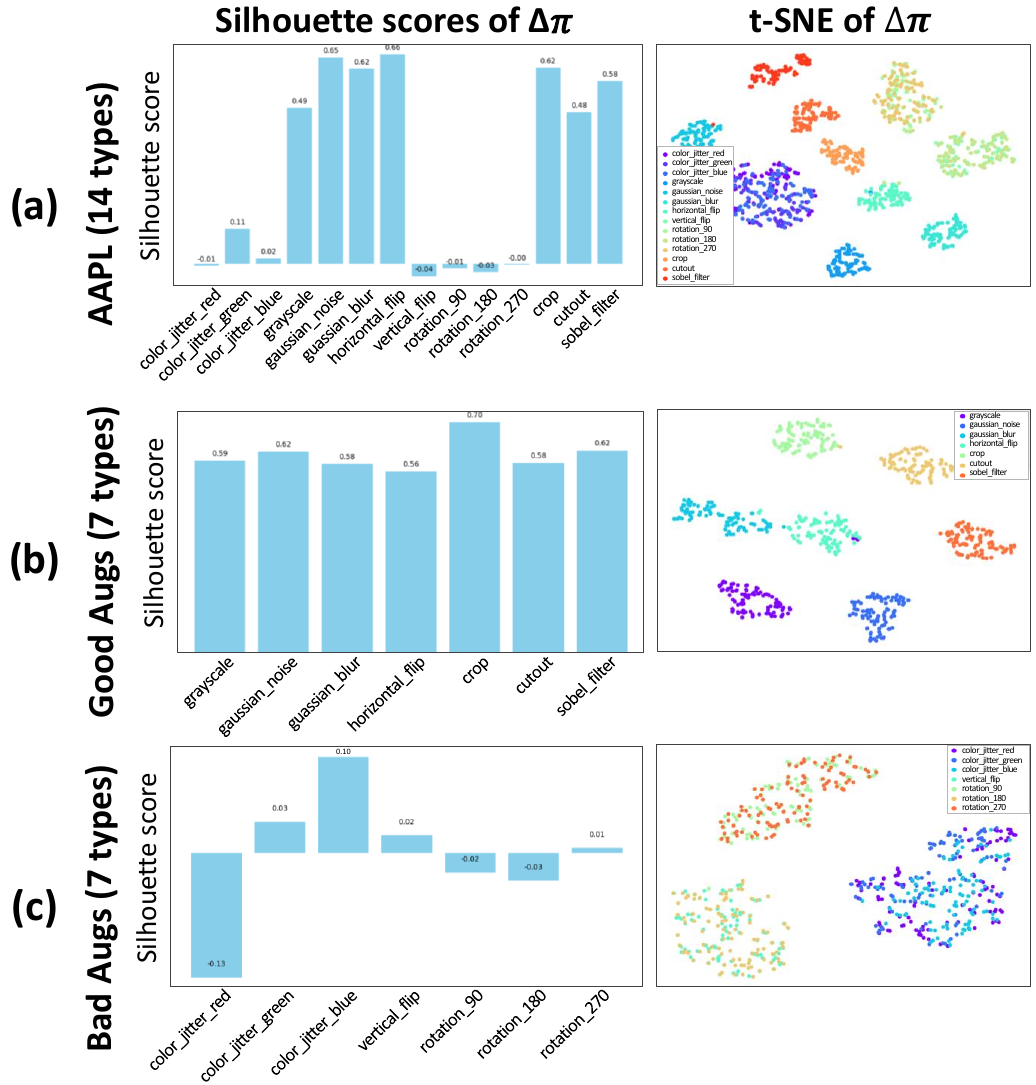}
    \caption{\textbf{The comparison of silhouette score and t-SNE} of the base-to-new generalization for each of the specific augmentation types on FGVCAircraft. All results are from the last epoch.}
    \label{fig:good_bad_aug}
    \vspace{-2mm}
\end{figure}

\noindent\textbf{Which augmentation is effective to prompt learning?}
~~~~The t-SNE visualization of the \textit{delta meta token} for 14 augmentations is shown, along with their silhouette scores, in Fig.~\ref{fig:good_bad_aug} (a). It turned out that it is difficult to distinguish rotations from flips and between color jitters, while other augmentations are obvious. All datasets exhibit difficulty in distinguishing these augmentations.
Following selective augmentation training, when trained only on augmentations whose results are good (shown in Fig.~\ref{fig:good_bad_aug} (b)), clustering is greatly enhanced, and silhouette scores are also raised. Also, the average performance for base-to-new generalization improved, as seen in Table~\ref{table:good_bad_aug}. But when training solely with the opposite, \ie, bad augs (Fig.~\ref{fig:good_bad_aug} (c)), there is neither significant improvement in silhouette scores nor in the average base-to-new generalization results. 
The ambiguity of augmentations between flips and rotations and between color jitters limits the learning capacity of the \textit{metanet}. 
\\

\begin{table}[t!]
\centering
\scriptsize
\begin{adjustbox}{width=0.8\linewidth}
\begin{tabular}{l|c|c|c}
\toprule[1pt]
Method & \texttt{\textbf{AAPL}} & Good Augs & Bad Augs  \\
\midrule
ImageNet & \textbf{73.09 }& 72.91 & 73.05  \\
Caltech101 & 95.87 & \textbf{96.43} & 96.00 \\
OxfordPets & 96.44 & \textbf{96.49} & 95.96  \\
StanfordCars & 71.70 & \textbf{71.85} & 71.67  \\
Flowers102 & \textbf{82.09} & 80.80 & 81.74  \\
Food101 & \textbf{91.10} & 90.45 & 90.90  \\
FGVCAircraft & \textbf{34.27} & 34.02 & 18.14 \\
SUN397 & 77.60 & 77.97 &\textbf{ 78.03}  \\
DTD & 60.31 & 61.24 & \textbf{61.43}  \\
EuroSAT & 64.15 & 66.68 & \textbf{74.70}\\
Ucf101 & 78.10 & 77.09 & \textbf{78.11} \\
\midrule
\textit{Average} & 74.97 & \textbf{75.08} & 74.52  \\
\bottomrule[1pt]
\end{tabular}
\end{adjustbox}
\caption{\textbf{\texttt{AAPL} with some augmentation types.} The comparison of harmonic means of base-to-new generalization accuracy when conducting \texttt{AAPL} using only good augmentations and bad augmentations.}
\label{table:good_bad_aug}
\vspace{-2mm}
\end{table}

\noindent\textbf{\texttt{AAPL} with weighted random sampling}~~~~
Fig.~\ref{fig:fig_sil_score} shows a consistent correlation between lower silhouette scores and worse zero-shot classification performance compared to CoCoOp~\cite{zhou2022conditional} across several datasets. Insufficient knowledge of semantic features makes classifying unseen classes more difficult. To address this, an active approach~\cite{tamkin2022active, ranganathan2017deep, konyushkova2017learning} was utilized for datasets DTD~\cite{cimpoi2014describing}, EuroSAT~\cite{helber2019eurosat}, StanfordCars~\cite{krause20133d}, and SUN397~\cite{xiao2010sun}, which have insufficient learning of augmentation type information. 
For training, silhouette scores were used as thresholds for random sampling weights. As shown in Table~\ref{table:wrs_ablation}, this improved the performance of base-to-new generalization across all 4 datasets. 
Notably, EuroSAT showed a significant 10\% improvement, emphasizing the effectiveness of dynamically selecting and emphasizing weaker augmentation types during each epoch. 
It demonstrates that attribute-specific feature decomposition for challenging augmentations enables more robust learning of semantic features.

\begin{table}[t]
\centering
\footnotesize
\begin{tabular}{lccg}
\toprule[1pt]
 & \texttt{\textbf{AAPL}} & WRS & $\Delta$ \\
\midrule 
StanfordCars & 71.70 & 71.82 & \textbf{+0.12} \\
SUN397 & 77.60 & 78.14 & \textbf{+0.54} \\
DTD & 60.31 & 61.39 & \textbf{+1.08} \\
EuroSAT & 64.15 & 74.25 & \textbf{+10.10} \\
\bottomrule[1pt]
\end{tabular}
\caption{\textbf{\texttt{AAPL} with weighted random sampling for vulnerable 4 datasets.} The comparison of harmonic means of base-to-new generalization accuracy. WRS is short for weighted random sampled \texttt{AAPL}. }
\label{table:wrs_ablation}
\vspace{-3mm}
\end{table}

\section{Conclusion}

Our novel approach efficiently extracts specific semantic features and delta meta tokens by subtracting the augmented image feature from the original image feature. Leveraging AdTriplet loss adversarially enhances classification loss, enabling precise discernment of attribute features through augmentations—a foundational aspect of our approach. By decomposing attribute and semantic features more accurately, we introduce attribute-specific bias into the prompt. Furthermore, our study underscores the indispensability of \texttt{AAPL} in prompt learning with augmentation for zero-shot classification tasks. In summary, our emphasis on attribute decomposition in prompt learning is underscored through augmentation profiling and analysis of dataset correlations, augmentations, and \texttt{AAPL} performance.
\\

\noindent\textbf{Acknowledgments}~~~~
Thanks to Prof. George Kamenos for his invaluable assistance in reviewing and editing this paper.
This work was partly supported by Innovative Human Resource Development for Local Intellectualization program through the Institute of Information \& Communications Technology Planning \& Evaluation(IITP) grant funded by the Korea government(MSIT) (IITP-2024-00156287, 40\%). 
This research was supported by the Korea Institute for Advancement of Technology (KIAT) grant funded by the Ministry of Trade, Industry, and Energy (MOTIE), Korea, (P0025331, 30\%). 
This work was supported by the National Research Foundation of Korea(NRF) grant funded by the Korea government(MSIT) (No. RS-2023-00252616, 30\%). \\

\label{sec:conclusion}

{
    \small
    \bibliographystyle{ieeenat_fullname}
    \bibliography{main}
}

\end{document}